\documentclass[conference]{IEEEtran}
\IEEEoverridecommandlockouts

\usepackage{cite}
\usepackage{mathtools}
\usepackage{amsmath,amssymb,amsfonts}
\usepackage{graphicx}
\usepackage{textcomp}
\usepackage{xcolor}
\usepackage{amsmath}
\usepackage{multirow}
\usepackage{multicol}
\usepackage{caption}
\usepackage{array}
\usepackage{subcaption}
\usepackage{balance}
\usepackage{wrapfig}
\usepackage{hyperref}
\usepackage{multirow}
\usepackage{multicol}
\usepackage{color}
\usepackage{parskip}
\usepackage{colortbl}
\usepackage{booktabs}
\usepackage{flushend}
\usepackage{hyperref}
\usepackage{todonotes}
\usepackage{comment}

\usepackage{cite}

\usepackage[utf8]{inputenc}
\usepackage[english]{babel}
\usepackage{enumitem}
\usepackage{algpseudocode}
\usepackage{algorithm}

\begin{document}

\title{Exploring Language Patterns in a Medical Licensure Exam Item Bank\

\thanks{This work was done while Swati was interning at NBME (Summer 2021).}
}

\author{\IEEEauthorblockN{Swati Padhee}
\IEEEauthorblockA{\textit{Department of Computer Science} \\
\textit{Wright State University}\\
Dayton, OH \\
\textit{padhee.2@wright.edu}}
\and
\IEEEauthorblockN{Kimberly Swygert, PhD, FCPP}
\IEEEauthorblockA{\textit{Test Development Services} \\
\textit{NBME}\\
Philadelphia, PA\\
\textit{kswygert@nbme.org}}

\and
\IEEEauthorblockN{Ian Micir}
\IEEEauthorblockA{\textit{Test Development Services} \\
\textit{NBME}\\
Philadelphia, PA\\
\textit{imicir@nbme.org}\\
 \\
}
}

\maketitle

\begin{abstract}
This study examines the use of natural language processing (NLP) models to evaluate whether language patterns used by item writers in a medical licensure exam might contain evidence of biased or stereotypical language. This type of bias in item language choices can be particularly impactful for items in a medical licensure assessment, as it could pose a threat to content validity and defensibility of test score validity evidence. To the best of our knowledge, this is the first attempt using machine learning (ML) and NLP to explore language bias on a large item bank. Using a prediction algorithm trained on clusters of similar item stems, we demonstrate that our approach can be used to review large item banks for potential biased language or stereotypical patient characteristics in clinical science vignettes. The findings may guide the development of methods to address stereotypical language patterns found in test items and enable an efficient updating of those items, if needed, to reflect contemporary norms, thereby improving the evidence to support the validity of the test scores. 

\end{abstract}


\section{Introduction}
Medical education has experienced a plethora of new challenges which could pose a significant social impact on the practice of medicine. Society places an unspoken moral contract on the profession of physicians, and the concept of medicine as both art and science further emphasizes the subjective nuances of care delivery \cite{cruess2008expectations}. Within the context of healthcare, the concept of prejudice – defined here as preconceptions regarding a patient based upon characteristics such as age, gender, ethnicity, or sexual orientation, when these aspects are \textit{not} relevant to patient care – is considered contrary to effective and empathetic medical practice and is discouraged within the medical profession. However, all humans can harbor unintentional biases based on their personal experiences. Such unintentional or implicit bias, defined as associations that exist subconsciously, can be subtle and challenging to detect \cite{fitzgerald2017implicit}. Some patient characteristics in medical decision-making have been historically marginalized, as the physicians’ primary focus is typically on the disease and symptoms of a given case. A focus on these patient characteristics is beginning to emerge in the literature; for example, a recent review of research in cardiovascular health highlights differences in the clinical presentation and subsequent diagnosis, and reveals the existent gender gap in patient outcomes in sophisticated modern health systems \cite{maas2010gender,langabeer2018sex}.

\begin{table*}[htbp]
\centering 
\caption{Example original practice stem before and after data pre-processing.}
 \scalebox{0.8}{
\vline
\begin{tabular}{| m{1.4\columnwidth} |m{1.0\columnwidth}|}
\hline
 \textit{Original Item Stem}  &\textit{Clean Item Stem}  \\
    \hline
    A 67-year-old woman with congenital bicuspid aortic valve is admitted to the hospital because of a 2-day history of fever and chills. Current medication is lisinopril. Temperature is 38.0°C (100.4°F), pulse is 90/min, respirations are 20/min, and blood pressure is 110/70 mm Hg. Cardiac examination shows a grade 3/6 systolic murmur that is best heard over the second right intercostal space. Blood culture grows viridans streptococci susceptible to penicillin. In addition to penicillin, an antibiotic synergistic to penicillin is administered that may help shorten the duration of this patient's drug treatment. Which of the following is the most likely mechanism of action of this additional antibiotic on bacteria?
    & 
    congenital bicuspid aortic valve admit hospital history chill current medication lisinopril temperature pulse respiration blood pressure cardiac examination grade systolic murmur well hear second right intercostal space blood culture grow viridan streptococci susceptible penicillin addition penicillin antibiotic synergistic penicillin administer help shorten duration drug treatment following likely mechanism action additional antibiotic bacteria\\
    \hline
\end{tabular}
}
\label{example}
\end{table*}

Physicians often serve as item\footnote{The term \textit{item} is used to describe a test question in any format.} writers for medical licensure exams, bringing to bear their individual experiences, which includes the training received in medical school, residency, internships, and other phases of medical education, as well as experiences in the workplace \cite{hirsh2014influence}. One very common item format for these exams is the patient vignette. These vignettes often feature a lengthy description of a patient, including characteristics such as age, gender, ethnicity, and so on, as a way of adding authenticity to the task of interpreting patient data. The vignette provides the history, physical, and laboratory findings, while the item stem is the specific question (e.g., \textit{Which of the following is the most likely diagnosis for this patient?}), and the item options contain the key and incorrect distractors \cite{nbmeitem}. When these items are coded for editing and test construction, the item metadata are likely to represent only the subset of these patient vignette factors that are relevant to the assessment point, such as diagnosis, gender, and age. Typically, the test construction process uses a relatively small set of codes to optimize the simultaneous alignment of content, statistical, and security related constraints. Thus, an item bank may be easily searchable for content related to age, or gender balance may be used as a constraint in form construction, while other patient characteristics go uncoded. 

As an item bank grows over time, unintended patterns of language related to these uncoded patient characteristics may emerge in the items. Each individual item within a test construction item bank may be of high quality and pass the quality control reviews, but examinees may see sets of items on test forms that feel like stereotypical or unrealistic representations of patients that appear to present a biased picture of what constitutes good medical practice. To address this, it would be useful to be able to rapidly quantify and evaluate patient characteristic language across large item banks, as this would support subsequent efforts to update item language across the item bank to reflect continuous improvement and alignment of language changes within society and medicine. 

\begin{figure}[htbp]
\includegraphics[width=1\columnwidth]{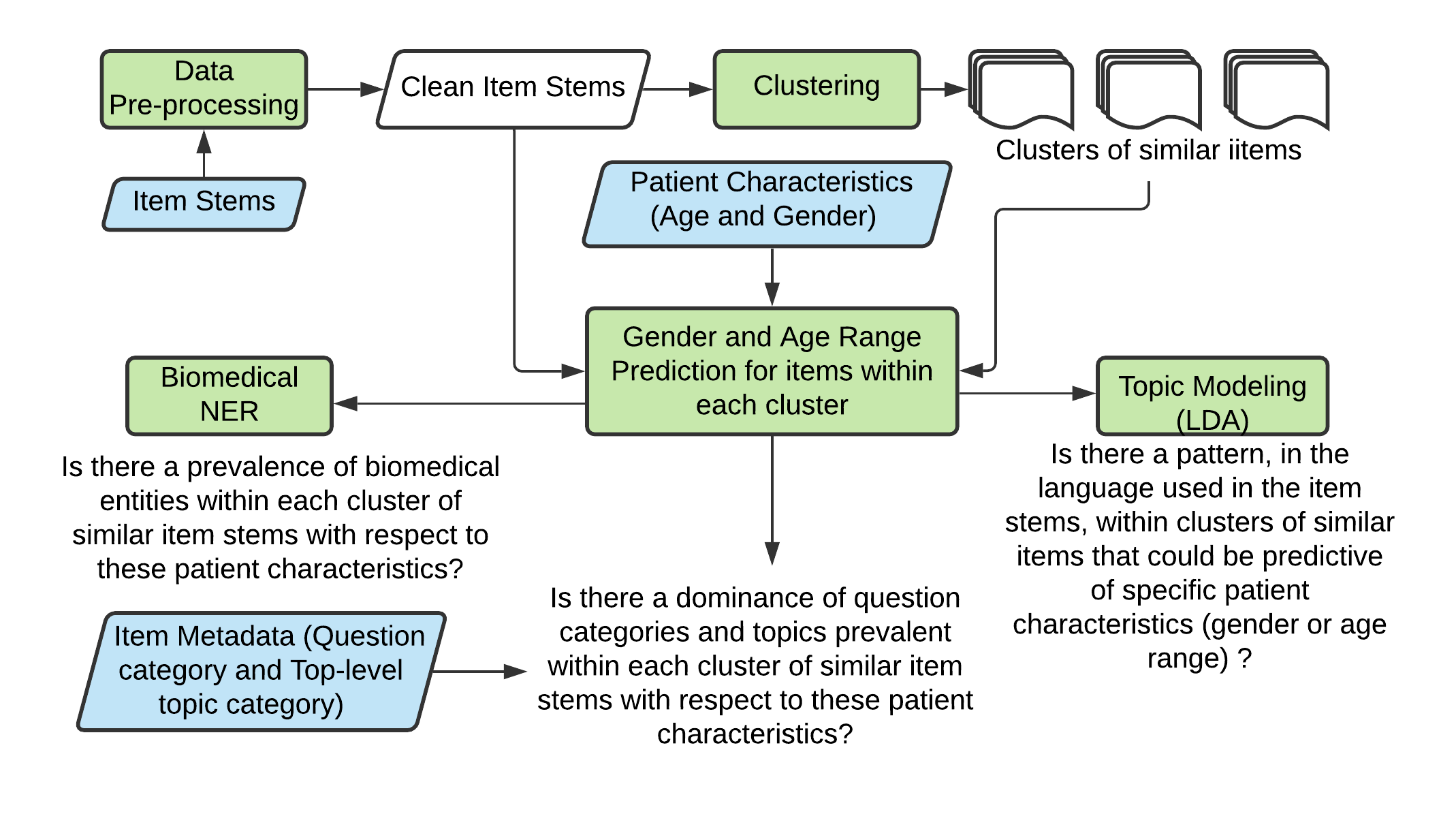}
\caption{Overview of the Methodology. The approach to address three research questions is shown. (Blue boxes show the data and green boxes show the tasks.)}
\label{fig:method}
\end{figure}

To date, much of the research related to exploring implicit bias occurring within healthcare has focused on either the gender of the patient \cite{woodward2019cardiovascular} or the physician \cite{bernardes2013engendering,sabin2009physicians}. In this study, we propose to explore such bias on the part of physicians writing items for medical licensure exams, where ``bias” is defined as the use of repeated, stereotypical language patterns with respect to patient characteristics. We defined three research questions (RQ) that would help evaluate language patterns in such a way to reveal this bias:
\begin{enumerate}
    \item Is there a pattern, in the language used in the item stems, within clusters of similar items that could be predictive of specific patient characteristics (gender or age range)?
    \item Is there a prevalence of biomedical entities within each cluster of similar item stems with respect to these patient characteristics?
    \item Is there a dominance of question categories and topics prevalent within each cluster of similar item stems with respect to these patient characteristics? 
\end{enumerate}

\section{Related Work}
This section discusses related work on exploring the usage of NLP in language bias detection. In the field of psychometrics, item bias research usually starts with the observation that a group membership is associated with the item responses, e.g,, the mean scores on items measuring vocabulary are higher for a particular examinee subgroup than for others, after controlling for overall ability on the test. However, any association with a group and item is not sufficient evidence that an item is biased towards that group \cite{mellenbergh1989item}. In psychology, the Implicit Association Test (IAT) is used to measure subconscious gender bias in humans, which can be quantified as the difference in time and accuracy for humans to categorize words as relating to two concepts they find similar versus two concepts they find different \cite{caliskan2017semantics,greenwald1998measuring}. In a healthcare setting, studies have shown that implicit bias may be observed through the actions of physicians as well as patients  \cite{gonzalez2014implicit,
matthews2002qualitative,amoli2016gender}. We propose to extend the research in bias to identify any language patterns that may emerge in large groups of items, such as item banks, when exhibited by item writers who may all be tailoring items towards the most common patient scenarios from their experience and individual reasoning. 

\section{Methodology}
We present an overview of our methodology in Figure \ref{fig:method}. For this study, we considered a large subset of items (N $>$ 20,000) from an item bank used for large-scale, standardized computer-based medical licensure assessments in the US\footnote{The data cannot be made available due to exam security considerations.}. These exams are administered only in English. All items were multiple-choice questions (MCQs) written by physician experts who had been trained to use a consistent and comprehensive set of style guidelines for creating patient vignettes within specific content areas in the foundational and clinical sciences. Every vignette’s text contained a reference to the patient's age and gender, consistent with the exam style guidelines, and every item contained metadata with a categorical value representing the age (grouped into categories) and gender. A sample item stem, representing a retired item\footnote{https://www.usmle.org/practice-materials/index.html}, is given in Table \ref{example}. Item stem here indicates the main body of an item, excluding the answer options. Each item is constructed to assess one of the competencies (physician task categories) and top-level topics (medical content areas).

\begin{table}[htbp]
\centering
\smallskip
\caption{Silhoutte Score for k = 2 to k = 7}
\scalebox{1.3}{
\begin{tabular}{|c|c||c|c|}
\hline
\textit{k} &\textit{Score} &\textit{k} &\textit{Score} \\
\hline
2 & 0.358 &5 & 0.412\\
\hline
3 & 0.385 &6 & 0.404\\
\hline
4 & 0.405 &7 & 0.393\\
\hline
\end{tabular}%
}
\label{tab:silhoutte}
\end{table}

\begin{figure}[htbp]
\centering
\includegraphics[width=0.7\columnwidth]{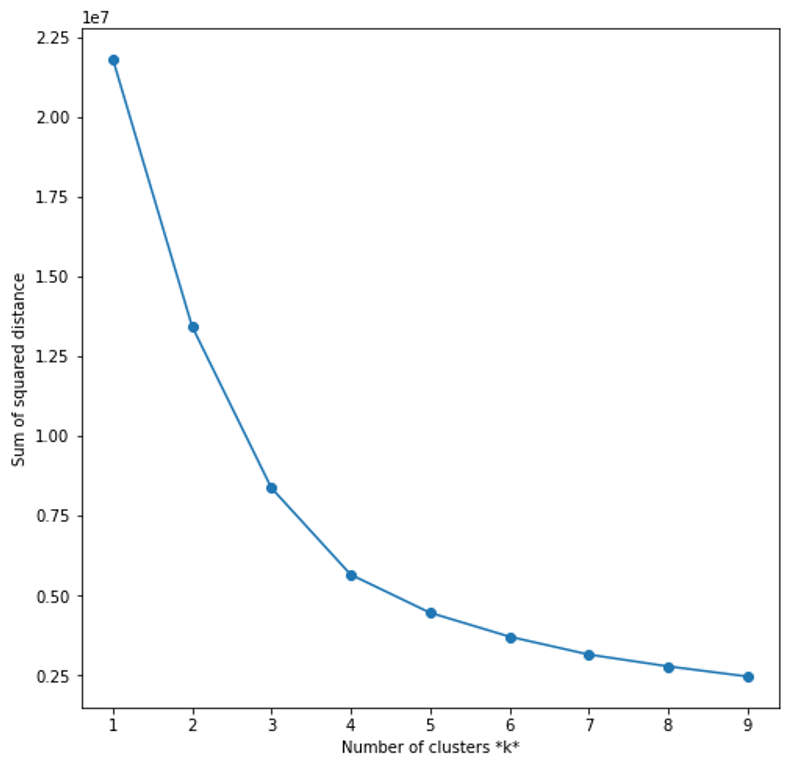}
\caption{Elbow Method showing Sum of squared distance for k = 1 to k = 9.}
\label{fig:elbow}
\end{figure}

\subsection{Data Pre-processing}
The item stems were first pre-processed, as discussed in this section. To preserve the context and semantics of the content, we first converted the item stems to lowercase and then removed the stop words, punctuations, symbols (\#,@), as well as special characters. We preserved the negation words as they are significant in the medical domain. Next, we removed the units (mg, ml, etc.) and numeric tokens as multiple items might have similar concepts with varying units and numeric values. As we intended to identify both patient characteristics indicative of language patterns, we removed age and gender indicative terms. We also removed high-frequency non-medical words after consultation with an internal medical advisor who has expertise in the content areas represented by the exam blueprints, in the item-writing guidelines, and in training other physicians to create content for these exams. Finally, we performed lemmatization and stemming to generate the pre-processed (clean) item stems as shown in Table \ref{example}. 

\subsection{Clustering}
We performed unsupervised clustering on the clean item stems as it would enable us to group items with similar content and then identify any prevalence of language patterns within similar items. We used K-means algorithm\cite{lloyd1982least} on embeddings generated for the clean item stems using SciBERT\cite{beltagy2019scibert} which is a pre-trained language model based on BERT\cite{devlin2018bert} trained on a large corpus of scientific text (1.14M papers from Semantic Scholar). We chose SciBERT representations for clustering as an error analysis on the items showed that a tokenizer built on SciBERT embeddings extracted more medical tokens such as ``diagnose" and ``diabetes mellitus" as compared to BERT or BioClinicalBERT\cite{alsentzer-etal-2019-publicly}. BioClinicalBERT model is is trained on all notes (approximately 880M words) from MIMIC III\footnote{https://mimic.mit.edu}, a database containing electronic health records from ICU patients at the Beth Israel Hospital in Boston, MA. The K-means clustering algorithm tries to make intra-cluster items as similar as possible while keeping the clusters as different as possible. To decide the optimal number of clusters, we first used the elbow method\footnote{https://en.wikipedia.org/wiki/Elbow\_method\_(clustering)}, which runs the clustering for a range of the number of K and for each value of K computes an average score for all clusters. The elbow or cutoff point indicates where we should choose a number of clusters, since it indicates that adding another cluster will not result in better modeling of the data. However, as shown in Figure \ref{fig:elbow}, there is no clear distinction between the number of clusters or elbow point. Hence, we performed a Silhouette analysis \cite{rousseeuw1987silhouettes}, which studies the separation distance between the resulting clusters. It measures how close each point in one cluster is to points in the neighboring clusters (the higher, the better). With the highest score of 0.41, as shown in Table \ref{tab:silhoutte}, we proceed with five clusters as the optimal number of clusters for our data. 
\begin{table}[htbp]
\centering
\smallskip
\caption{Gender prediction results (accuracy) (F:Female, M:Male)}
\scalebox{0.8}{
\begin{tabular}{|c|c|c|c|c|c|}
\hline
\textit{Cluster} &\textit{Baseline (F)} &\textit{Baseline (M)} &\textit{TF-IDF} &\textit{SciBERT} &\textit{BioClinicalBERT}\\
\hline
1 &0.48 &0.52 & 0.64 &0.60 & 0.38 \\
\hline
2 &0.44 &0.56 & 0.67 & 0.65 &0.30 \\
\hline
3 &0.74 &0.26 & 0.84 & 0.79 &0.78 \\
\hline
4 &0.48 &0.52 & 0.63 & 0.63 &0.60 \\
\hline
5 &0.49 &0.51 & 0.68 & 0.60 &0.61 \\
\hline
\end{tabular}%
}
\label{tab:genTF}
\end{table}

\begin{table}[htbp]
\centering
\smallskip
\caption{Age prediction results (accuracy)}
\scalebox{0.9}{
\begin{tabular}{|c|c|c|c|c|}
\hline
\textit{Cluster} &\textit{Baseline Average} &\textit{TF-IDF} &\textit{SciBERT} &\textit{BioClinicalBERT}\\
\hline
1 & 0.17 & 0.40 &0.33 & 0.32 \\
\hline
2 &0.17 &0.50 & 0.45 & 0.39 \\
\hline
3 &0.25 &0.81 & 0.76 & 0.78 \\
\hline
4 &0.17 &0.50 & 0.41 & 0.42 \\
\hline
5 &0.20 &0.58 & 0.49 & 0.50 \\
\hline
\end{tabular}%
}
\label{tab:ageTF}
\end{table}

\subsection{Gender and Age Range prediction}
Once we grouped our item stems into five different clusters, we performed gender and age prediction within each cluster. We have the actual age and gender information for each item in a cluster from item metadata (patient characteristics). For age prediction, we defined seven age groups as advised by the internal clinical advisor: 0-5 years, 6-17 years, 18-34 years, 35-49 years, 50-64 years, 65-84 years, and 85+ years. Next, we designed one gender prediction task and one age range prediction task within each of the five clusters. For each of the ten tasks, we trained four traditional ML algorithms, i.e., Linear Support Vector Classification (LinearSVC), Logistic Regression (LogR), Multinomial Naıve Bayes (MNB), and Random Forest classifier (RFC) on context-free Term Frequency Inverse Document Frequency (TF-IDF) text vectorization representation. TF-IDF is a statistical measure that evaluates how relevant a word is to an item in a collection of items. We also evaluated each of the prediction tasks using two fine-tuned state-of-the-art contextualized language models for a medical domain to capture semantic relatedness in biomedical concepts, i.e., SciBERT \cite{beltagy2019scibert} and BioClinicalBERT\cite{alsentzer-etal-2019-publicly}. We took a conscious decision to evaluate our model performance on raw data instead of utilizing sampling methods to address data imbalance across clusters, as the scope of this study was to explore NLP techniques on raw data. In the future, we plan to extend our experiments with re-sampling methods. 
 
We randomly shuffled and split each cluster data into 80\% training data and 20\% test data. We created a gender prediction and an age range prediction model within each cluster on the training data to predict the test data's unknown gender and age range. Summarized, the input to our prediction algorithm is the item stem, and the unknown output it generates on the test set is the age range or gender. We then evaluated the performance of our models by comparing the generated predictions on the test set with the truth values that were hidden from the model.

\subsection{Topic Modeling - RQ 1}
Topic modeling provides methods for automatically organizing, understanding, searching, and summarizing documents. We explored topic modeling on the clean item stems with correctly predicted gender and age ranges to discover the hidden themes (language patterns indicative of specific patient characteristics) in a cluster. We utilized Latent Dirichlet Allocation (LDA)  \cite{Blei2003-qr} which is an unsupervised probabilistic model to automatically identify the hidden themes (topics) in a group of documents. It classifies the documents (i.e., item stems) into topics that best represent the data set (cluster). For the purpose of illustration, we limited our modeling to two major topics within each prediction.  

\subsection{Biomedical Named Entity Recognition - RQ 2}
In order to get further assistance in understanding the hidden language patterns in the clusters of similar items, we performed biomedical named entity recognition (NER) on the prediction results. We used a scispaCy Python package\footnote{https://allenai.github.io/scispacy/} containing spaCy models for processing biomedical, scientific, or clinical text. The NER model we use was trained on the BIONLP13CG corpus to extract entities with an F1 measure of 77.75 \% belonging to different entity types. 

\subsection{Item Metadata Analysis - RQ 3}
We compared the number of items correctly predicted in each of the ten prediction tasks across all five clusters to understand whether there was a prevalence of a particular question or topic category\footnote{https://www.usmle.org/prepare-your-exam/step-1-materials/step-1-content-outline-and-specifications} in the correctly predicted items (thereby the ones indicating bias towards patient characteristics).

\section{Results}

\subsection{Gender and Age Range Prediction}

We show the results in terms of accuracy in the gender-prediction task on all five clusters in Table \ref{tab:genTF}. Baseline accuracy for Female and Male was calculated based on their item distribution within each cluster using the below equation:

\begin{equation}
Baseline\: Accuracy_{i} = \frac{N_{i}}{ \sum_{1}^{n} N_{i}}
\end{equation}
where $N_{i}$ = number of items for category \textit{i} in a cluster and \textit{n} is the number of categories.

The accuracies for models trained on TF-IDF, SciBERT, and BioClinicalBERT representations were calculated using Scikit-learn package \cite{scikit-learn}. As accuracy here indicates the patterns related to gender patient characteristics, the ideal outcome is for a model to have a lower accuracy, which would indicate that undesired patterns are not emerging in the item data. 

As Logistic Regression performed the best in all clusters, we select it for comparison with models trained on other representations as shown in Table \ref{tab:genTF}. 
Language models provide context to distinguish between words and phrases that sound similar. Hence, we can see that although TF-IDF models might capture repetitive patterns, contextualized models help us identify patterns of terms used in different contexts, thereby \textit{reducing} the accuracy. The variation in accuracy across different clusters is due to the diverse data distribution. Furthermore, as SciBERT and BioClinicalBERT models are built on different datasets, we observed a varying scale of performance across clusters depending on the nature of concepts in the items clustered together. We plan to do further in-depth data analysis on this varying performance in the future.

We show the results in terms of accuracy within the age prediction task on all five clusters in Table \ref{tab:ageTF}. Baseline average accuracy for all age groups is calculated based on their item distribution within each cluster as shown in Equation 1 and then averaged across age groups. Similar to gender, as accuracy here indicates the patterns related to age range patient characteristics, ideally, a model should have lower accuracy. As Logistic Regression performed the best with TF-IDF representation in all the clusters, we select it for comparison with models trained on other representations as shown in Table \ref{tab:ageTF}. We observed similar results comparing different feature representations that although TF-IDF models might capture more repetitive patterns (higher accuracy), contextualized models help us identify patterns of terms used in different contexts, thereby \textit{reducing} the accuracy. As SciBERT and BioClinicalBERT models are trained on different datasets, we observed that while SciBERT performed with higher accuracy in two clusters, BioClinicalBERT performed best in three others. We believe it is due to the nature of concepts in the items clustered together as well as the distribution of items belonging to different ages within a cluster. We plan to further investigate in this direction in the future. 

\subsection{Topic Modeling}

For illustration purposes, we discuss top two topics for age and gender prediction results of one cluster (cluster 3). 
The most relevant terms representing two topics indicate a prevalence of indicative female terms such as vaginal, breast, pelvic, gravida, para, delivery, gestation, pregnancy, uterus, vital, menstrual, etc. Furthermore, the correctly-predicted items in the age group of 18-34 years old in the same cluster contain more female-representative terms such as pelvic, pain, gestation, prenatal, pregnancy, etc. This indicates a prevalence of female reproductive-indicative terms in clean item stems in cluster 3 for 18- to 34-year-old females.

\subsection{Biomedical Named Entity Recognition}
For illustration purpose, we present the most prevalent biomedical named entities extracted from cluster 3 and 5 for visualization purpose in multiple word-frequency clouds. The larger the size of a term indicates it's higher frequency.

\section{Discussion}
\begin{figure}[htbp]
\centering
\includegraphics[width=0.8\columnwidth]{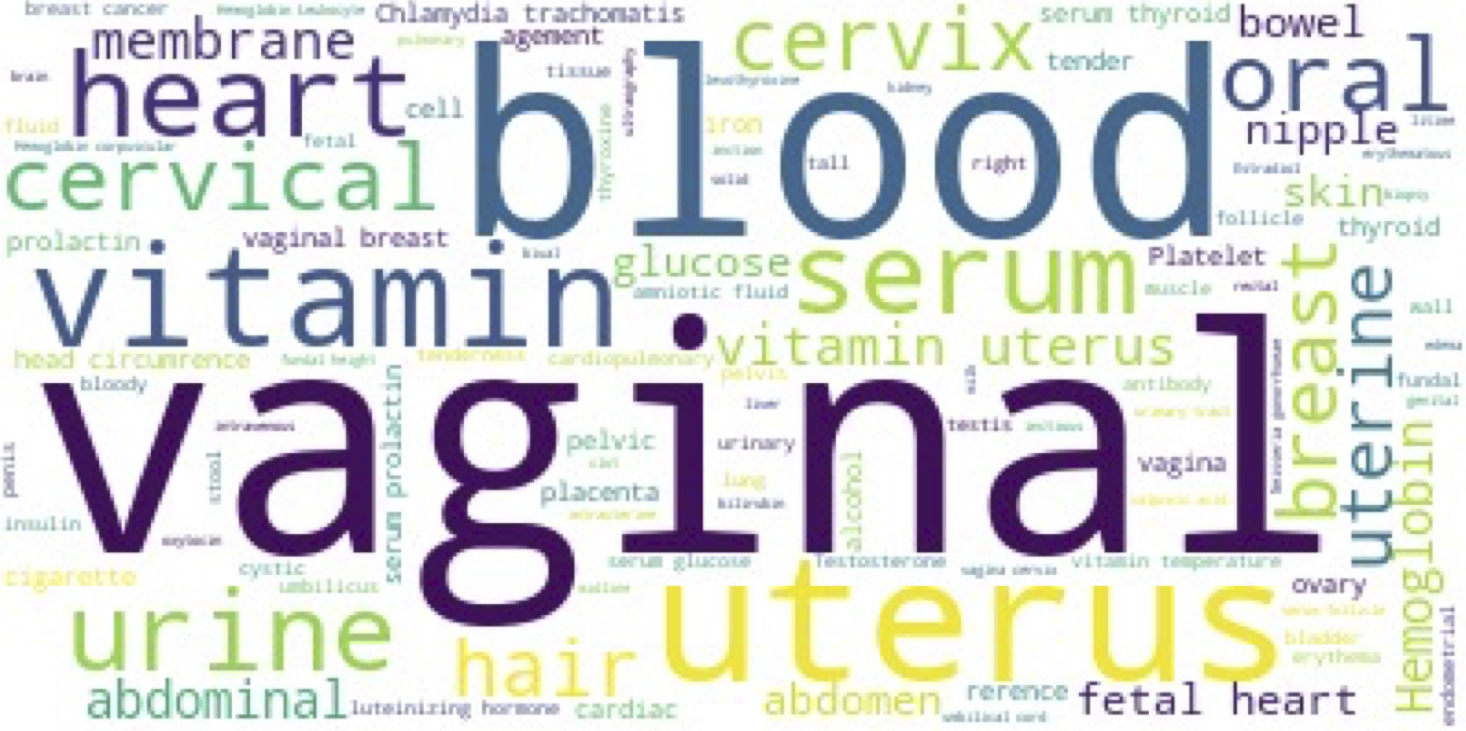}
\caption{Biomedical named entities prevalent in correctly predicted 18-34 years age range items in example cluster 3.}
\label{fig:bio18C3}
\end{figure}

\begin{figure}[htbp]
\centering
\includegraphics[width=0.8\columnwidth]{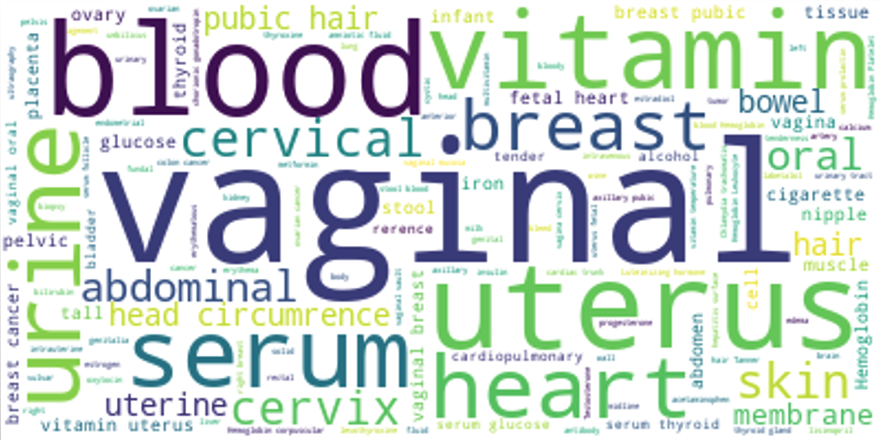}
\caption{Biomedical named entities prevalent in correctly predicted Female items in example cluster 3.}
\label{fig:bioFemaleC3}
\end{figure}

The results presented in the previous section led to the below findings.
We observed a prevalence of female reproductive-indicative terms in clean item stems in cluster 3 illustrated for 18- to 34-year-old females (Figures  \ref{fig:bio18C3},\ref{fig:bioFemaleC3}). 
    
\begin{figure}[htbp]
\centering
\includegraphics[width=0.8\columnwidth]{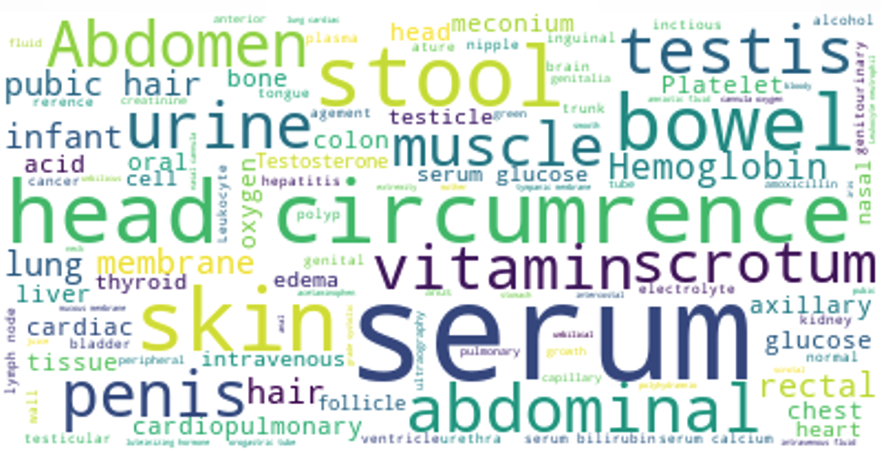}
\caption{Biomedical named entities prevalent in correctly predicted Male items in example cluster 3.}
\label{fig:bioMaleC3}
\end{figure}
    
    This is not necessarily unexpected or indicative of any bias, and in fact may be expected and intended by those responsible for content generation due to the requirements of the exam blueprint. However, this methodology provides a more subtle quantitative capture of the actual prevalence of such language in the item bank than could be obtained with a simple keyword search of the bank. Thus, these data could be very useful in motivating further discussions amongst advisory physicians and item writers to compare the findings with real-world expectations of prevalence of patient characteristics in certain clinical scenarios. Had unexpected language that is outdated or stereotypical emerged here, that would also be a basis for discussion, and again, this methodology would allow that language to emerge from the raw data, rather than from assumptions about the language used in the items.  
    
\begin{figure}[htbp]
\centering
\includegraphics[width=0.8\columnwidth]{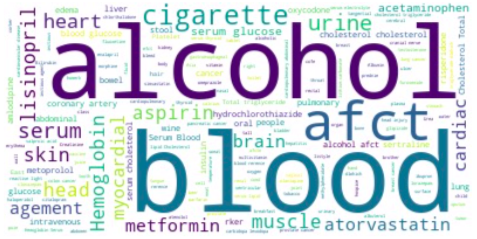}
\caption{Biomedical named entities prevalent in correctly predicted Male items in example cluster 5.}
\label{fig:bioFemaleC5}
\end{figure}

\begin{figure}[htbp]
\centering
\includegraphics[width=0.8\columnwidth]{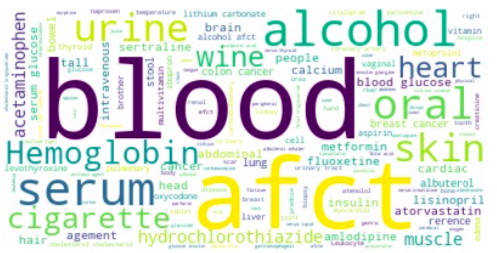}
\caption{Biomedical named entities prevalent in correctly predicted Female items in example cluster 5.}
\label{fig:bioMaleC5}
\end{figure}

We found that in addition to female- and male-indicative terms in the illustrated cluster 3, we identified new entities such as ``oral” emerging among 18- to 34-year-old females. Furthermore, we also found that in another cluster 5, other than the gender-indicative terms, while ``alcohol” was found to be more dominant among Male items (Figure \ref{fig:bioFemaleC5}), ``wine” was more dominant among Female items (Figure \ref{fig:bioMaleC5}).
    
    This is extremely useful data for the set of physicians who create content for this exam, as they can use it to discuss if this language is emerging due to an increasing focus on oral symptoms or disease among female patients in the 18- to 34-year old age group; whether it makes clinical sense that the dominance of language related to oral issues among females is larger (Figure \ref{fig:bioFemaleC3})  as compared to males (Figure \ref{fig:bioMaleC3}); and whether the prevalence in the item bank matches with real-world case rates.
    Similarly, physicians can discuss whether alcohol usage is emerging as a relevant concern among female patients, and whether references specifically to wine consumption is appropriately reflective of patient characteristics, or is stereotypical language that should be revised. 
    
Our item metadata analysis showed a uniform distribution across all clusters and all prediction tasks. For example, in the illustrative cluster 3 under discussion, we found the top-level topic of ``Reproductive \& Endocrine Systems" to be distributed equally across males, females, and all age groups. Similarly, the three major question categories with a uniform distribution were ``Medical Knowledge: Applying Foundational Science Concepts," ``Patient Care: Diagnosis," ``Patient Care: Management." This indicates that there is no significant prevalence of questions and topic categories within clusters - an ideal scenario.

The limitation of this study is that the analysis could be done on only two patient characteristics (i.e., age and gender) as these were the only two consistently present in the metadata. However, this study could be extended to other characteristics such as race and ethnicity to further investigate their correlation within the item bank and real-world case rates, for items which contain these specific metadata. Furthermore, using data-driven similarity-based clustering methods resulted in an imbalanced distribution of items across different age groups and gender within each cluster. Although our experiments on raw imbalanced data groups demonstrated that NLP and ML algorithms can be used to identify language patterns, extending this study in the future to re-sampling methods, as well as other approaches, may lead to a more in-depth understanding of relationship between the patterns identified with the patient characteristics. 

\section{Conclusion}
To the best of our knowledge, this study presents the first exploratory analysis of large item banks in a medical licensure exam to identify language patterns used by item writers. Our results demonstrate the feasibility of machine learning models trained on contextual representations to extract language patterns from large item banks which can be utilized to compare with the real-world case scenarios. The findings suggest that natural language processing can be used to review large item banks for potential stereotypical patient characteristics or language patterns in clinical science vignettes, where this would also have the potential to improve the defensibility of the item language and the evidence to support the use of such items in these exams.

\section*{Acknowledgment}
The authors are deeply grateful to Victoria Yaneva, PhD, and Miguel Paniagua, MD, FACP, FAAHPM, for their valuable feedback in this paper.

\bibliographystyle{IEEEtran}
\bibliography{IEEEabrv,sample}
\end{document}